\documentclass[10pt, conference]{IEEEtran}
\IEEEoverridecommandlockouts
\usepackage{cite}
\usepackage{natbib}
\usepackage{amsmath,amssymb,amsfonts}
\usepackage{graphicx}
\usepackage{subcaption}
\usepackage{textcomp}
\usepackage{xcolor}
\usepackage{amssymb, bm}
\usepackage{algorithm}
\usepackage{algpseudocode}
\usepackage{booktabs}
\usepackage{xcolor}
\usepackage{url}

\def\BibTeX{{\rm B\kern-.05em{\sc i\kern-.025em b}\kern-.08em
    T\kern-.1667em\lower.7ex\hbox{E}\kern-.125emX}}

\begin{document}

\title{Diff-Feat: Diffusion-Based Cross-Modal Feature Extraction for Multi-Label Classification}

\author{\IEEEauthorblockN{Tian Lan$^{1}$, Yiming Zheng$^{1}$, Jianxin Yin$^{1,\dagger}$}
\IEEEauthorblockA{\textit{$^{1}$ Center for Applied Statistics and School of Statistics, Renmin University of China, Beijing, China}}
\IEEEauthorblockA{\textit{$^{\dagger}$ Corresponding author: jyin@ruc.edu.cn}}

\thanks{%
© 2026 IEEE. This is the author's accepted manuscript.

T. Lan, Y. Zheng, and J. Yin,
``Diff-Feat: Diffusion-Based Cross-Modal Feature
Extraction for Multi-Label Classification,''
in \emph{2026 International Joint Conference
on Neural Networks (IJCNN)}, 2026.

Published version:
\protect\url{https://doi.org/10.1109/IJCNN69854.2026.11728464}
}
}

\maketitle

\begin{abstract}
 Multi-label classification has broad applications and depends on powerful representations capable of capturing multiple interactions. We introduce \textit{Diff-Feat}, a simple yet effective framework that extracts intermediate features from pre-trained diffusion-Transformer models for both images and text, and fuses them for downstream tasks. We observe distinct patterns in optimal feature selection: for vision tasks, the most discriminative intermediate feature along the diffusion process occurs at the middle step and is located at the middle block in Transformer. In contrast, for language tasks, the best feature occurs at the noise-free step and is located at the deepest block. Notably, a consistent phenomenon across diverse datasets: a fixed "Middle Layer" yields the best performance on various downstream classification tasks for images (under DiT-XL/2-256$\times$256 and various U-ViT architectures). We further analyze the underlying mechanism and its scaling behavior across model scales. To efficiently select representations, we devise a heuristic local-search algorithm that pinpoints the locally optimal "image-text"$\times$"block-timestep" pair among a few candidates, avoiding an exhaustive grid search.  A simple fusion-linear projection followed by addition-of the selected representations achieves 46.1\% mAP on Visual Genome 500 (with 500 categories) and 80.2\% mAP on Pix2Cap-COCO (with 80 categories), consistently outperforming strong baseline methods. Code is available at \url{https://github.com/lt-0123/Diff-Feat}.
\end{abstract}


\section{Introduction}

Multi-label classification has wide-ranging applications in image retrieval, biomedical image recognition~\citep{zongyuan2018chest}, and scene understanding~\citep{shao2015sceneunderstanding}. Compared with single-label classification, it poses two major challenges~\citep{Query2Label}: label imbalance and the difficulty of extracting features from regions of interest, which demands more expressive feature representations.

Recent advances in diffusion models~\citep{2015Deepthermodynamics, 2020DDPM, song2021scorebased} have demonstrated remarkable generative capabilities across multi-modal domains such as image synthesis~\citep{2021DiffusionbeatGAN, 2022text_to_image}, audio generation~\citep{kong2021diffwave, huang2023audio}, and natural language processing~\citep{li2022Diffusionlm, nie2025large}. Beyond generative modeling, diffusion models have recently attracted growing interest as representation learners for downstream tasks~\citep{baranchuk2022labelefficient, Xiang_2023_ICCV, Xu_2023_CVPR}, leveraging their denoising process to learn rich semantic features~\citep{baranchuk2022labelefficient}. The likelihood-based training objective gives the model strong potential to function as a foundation model in a self-supervised setting.

Furthermore, for language modeling, diffusion-based language models offer greater flexibility than conventional autoregressive models in capturing diverse language-order dependencies by supporting any-order generation~\citep{xue2025anyordergptmaskeddiffusion}, thereby opening new possibilities for learning more comprehensive and robust semantic representations. This property also makes diffusion language models a promising foundation for extracting high-quality representations for downstream analysis and understanding tasks.

\begin{figure}[t]
  \centering
  \includegraphics[width=0.5\textwidth]{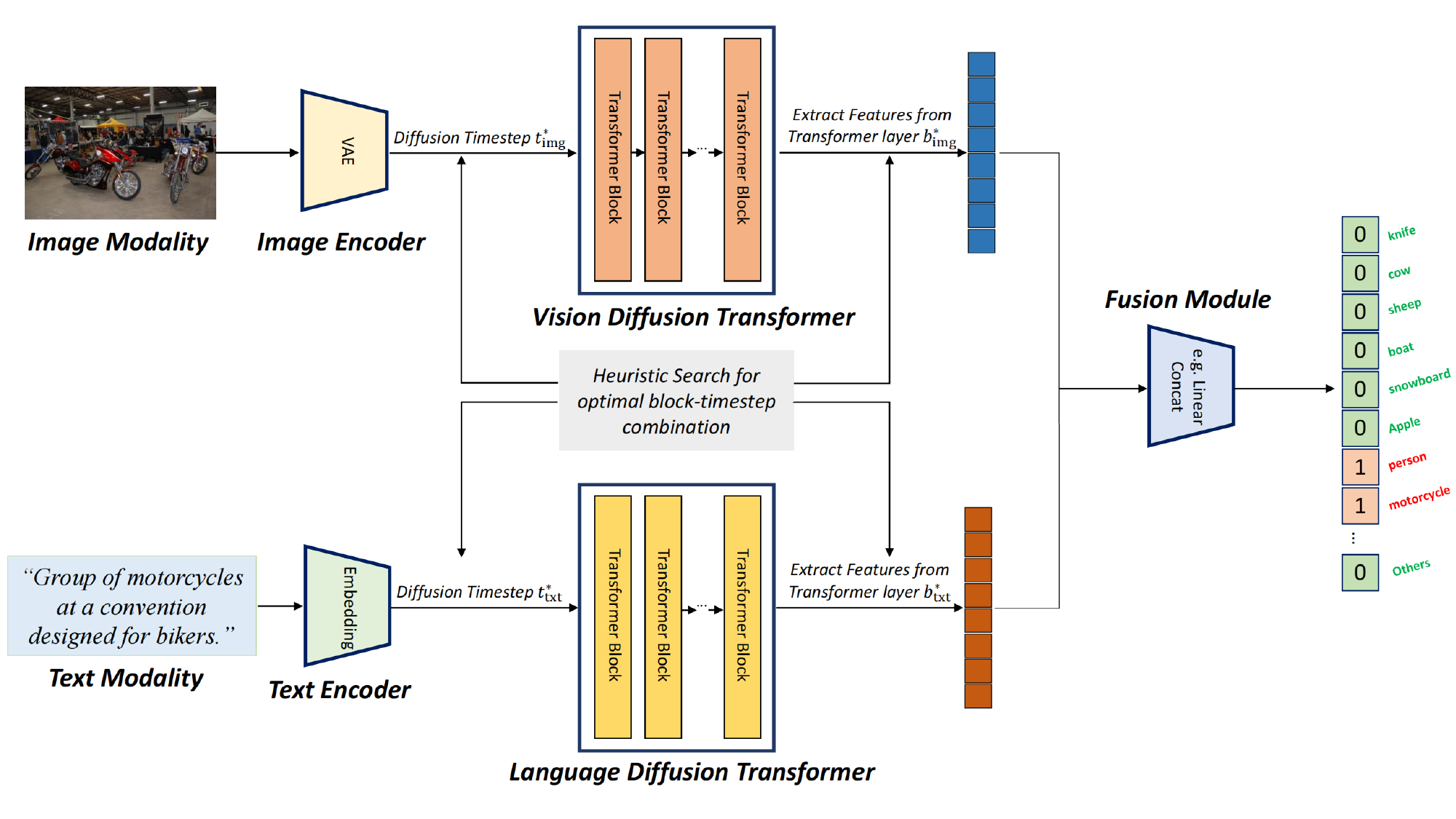}
  \caption{Overview of the Diff-Feat framework.}
  \label{fig:overview}
\end{figure}

Inspired by the strong linear separability and semantic understanding capabilities of diffusion-based representations~\citep{Xiang_2023_ICCV}, a central question arises: \textit{How can we identify the optimal pair of diffusion-based representations from image and text to enhance the performance of downstream multi-label classification?} To address this question, we propose  \textit{Diff-Feat}, a simple yet effective framework for multi-label classification. As shown in Fig.~\ref{fig:overview}, Diff-Feat extracts visual and textual features from pre-trained continuous diffusion-Transformer models across different noise levels and Transformer blocks, and then fuses them to perform multi-label prediction.

To the best of our knowledge, we are the first to treat image and text modalities symmetrically by independently extracting their diffusion representations. We conduct a comprehensive empirical study that characterizes the effectiveness of representations extracted at different noise levels and Transformer blocks across both modalities. Furthermore, we propose a simple heuristic-guided local search algorithm to efficiently identify the optimal image-text representation pair. When fused, the selected representations achieve leading performance on multi-label classification benchmarks: \(46.1\%\) mAP on Visual Genome 500~\citep{VGdataset} and \(80.2\%\) mAP on Pix2Cap-COCO~\citep{you2025pix2capcocoadvancingvisualcomprehension}.

Meanwhile, we observe a consistent and robust phenomenon: for image data, regardless of the diffusion timestep, dataset distribution, downstream classification task, evaluation metric, or model architecture and scale, the most discriminative features consistently emerge from the \textbf{middle} Transformer block of image diffusion-Transformer models. We further investigate the underlying mechanisms and examine how this pattern scales across various diffusion-Transformer architectures.

Our main contributions are summarized as follows:
\begin{itemize}
    \item We propose \textit{Diff-Feat}, a simple yet effective framework that extracts cross-modal diffusion representations for multi-label classification. Using a heuristic strategy to identify optimal fusion points, our method achieves strong performance on Visual Genome 500 and Pix2Cap-COCO.
    
    \item We present a large-scale and systematic empirical study that characterizes how Transformer depth and diffusion noise jointly shape representation quality across visual and textual modalities. Our analysis spans multiple datasets, diffusion timesteps, and probing tasks, providing principled insights into modality-specific robustness patterns and the emergence of discriminative semantics.
    
    \item We systematically uncover a robust and architecture-agnostic phenomenon: across diffusion timesteps, datasets, downstream tasks, evaluation metrics, model scales, and visual diffusion architectures, representations from the \textbf{middle} Transformer layers consistently yield the strongest discriminative performance. Through extensive scaling experiments and modality-specific mechanistic analyses, we show that this effect is tightly linked to model depth in vision models and driven by noise-signal trade-offs in vision diffusion systems.
    
\end{itemize}

\section{Background and related work}

\textbf{Multi-label classification.} Multi-label classification is a supervised learning task where an instance can be associated with multiple labels. In recent years, various deep learning techniques have been applied to address multi-label classification. Existing approaches explore diverse strategies for multi-label classification, including Transformer-based querying~\citep{Query2Label}, graph-based modeling~\citep{GKGNet}, and multimodal alignment~\citep{Xu2022OpenVM}. Meanwhile, cross-modal approaches~\citep{YuanChenYeBhattSaradeshmukhHossain+2023, WANG2022108002, ouyang2023, hoffmann2023, huangdiffdis} improve multi-label classification performance while effectively alleviating overfitting on majority classes. However, effectively addressing imbalanced label distributions and capturing complex label dependencies remains challenging and largely unresolved problems.

\textbf{Diffusion models and diffusion representations.} Diffusion models~\citep{2020DDPM} define a forward process in which an input \(\mathbf{x}_0\) is progressively corrupted by Gaussian noise over a series of timesteps \(t = 1, \cdots, T\). At each timestep, the noisy sample \(\mathbf{x}_t\) is sampled from the distribution \(q(\mathbf{x}_t|\mathbf{x}_0) = \mathcal{N}(\alpha_t \mathbf{x}_0, \sigma_t^2 \mathbf{I})\), where \(\alpha_t = \sqrt{\prod_{i=1}^t (1 - \beta_i)}\) and \(\alpha_t^2 + \sigma_t^2 = 1\). The noise schedule \(\beta_1, \ldots, \beta_T\) is determined by a linear schedule from \(\beta_{\min}\) to \(\beta_{\max}\), i.e.,
\begin{equation}
\label{eq:diffusion_forward}
    \mathbf{x}_t = \alpha_t\mathbf{x}_0 + \sigma_t \bm{\epsilon}
\end{equation}
Inspired by the representation power of Denoising Autoencoders (DAEs)~\citep{DAE2013, Vincent2008encoder} in compressed latent spaces, recent work~\citep{baranchuk2022labelefficient, Xiang_2023_ICCV, textfree2024,zhang2023} has increasingly explored the representation learning potential of diffusion models. However, existing research has primarily focused on images, with limited attention given to extracting representations from language-based diffusion models.

\section{Approach}
\label{sec: Approach}

\subsection{Discriminative diffusion representations for image and text}

Inspired by prior work~\citep{Xiang_2023_ICCV} that leverages intermediate activations from pre-trained image diffusion models, we adopt a similar philosophy. This strategy requires no modification to standard diffusion backbones and remains fully compatible with existing models.

Based on this idea, we also utilize intermediate activations extracted from pre-trained continuous language diffusion models~\citep{lovelace2023latent, plaid}, focusing on specific decoder blocks and noise levels. Unlike previous approaches~\citep{Xu_2023_CVPR, Li_2024_CVPR} that treat text as a conditional embedding \(\tau(\mathbf{s})\) and extract intermediate activations via \(f=\bm{\epsilon}_{\bm{\theta}}(\mathbf{x}_t, \tau(\mathbf{s}), t)\), where \(\tau(\mathbf{s})\) denotes the embedding of the image caption \(\mathbf{s}\) by a pre-trained text encoder \(\tau\) and \(\bm{\epsilon}_{\bm{\theta}}\) denotes the diffusion model parameterized by \(\bm{\theta}\), we treat both text and image as equal modalities for representation learning. This symmetric strategy provides greater flexibility for selecting task-relevant features from different layers and noise levels.

To apply noise, we randomly sample \(\bm{\epsilon} \sim \mathcal{N}(\mathbf{0}, \mathbf{I})\) and apply Eq.~\ref{eq:diffusion_forward} to obtain \(\mathbf{x}_t\) for images, since no significant differences are observed between random and deterministic noising methods~\citep{Xiang_2023_ICCV}. However, for text, we find that deterministic noising (e.g., DDIM~\citep{ddim}) yields better performance.

Formally, we define the problem as identifying the optimal diffusion timestep \( t \in \mathcal{T} \) and decoder block \( b \in \mathcal{B} \) that minimize the discriminative loss on a downstream task, i.e., \((t^*, b^*) = \arg\min_{t \in \mathcal{T}, b \in \mathcal{B}} \mathcal{L}(t, b)\), where \( \mathcal{L}(t, b) \) denotes the downstream discriminative loss, \(\mathcal{T}\) and \(\mathcal{B}\) denote the sets of diffusion timesteps and decoder blocks, respectively.

We conduct linear probing to identify diffusion representations with strong linear separability and label semantics. In addition, we fine-tune the image diffusion model to enhance downstream performance.

\subsection{Fusion strategy with unimodal diffusion representations}
\label{sec: representation fusion}

Furthermore, the performance of multi-label tasks can be significantly improved by selecting the optimal "image-text"$\times$"block-timestep" pairs and employing an effective fusion method. However, performing a grid search over all possible block-timestep combinations is computationally impractical, with a high complexity of \(O(|\mathcal{T}|^2|\mathcal{B}|^2)\), where \(|\cdot|\) denotes the cardinality of the set.

To address this challenge, we adopt a heuristic search (Algorithm~\ref{alg:heuristic_search}), where \(\mathrm{img}\) and \(\mathrm{txt}\) denote the image and text modalities, respectively. We first identify the optimal configuration for each modality, thereby reducing the search space by focusing on high-potential candidates. We then conduct a localized grid search within the neighborhoods of these unimodal optima to find the best fusion configuration. This approach significantly lowers computational cost while maintaining competitive performance, with a reduced complexity of \(O(|\mathcal{T}||\mathcal{B}|)\).

\begin{algorithm}[htbp]
\caption{Heuristic Local Search for Fusion Block-Timestep Selection}
\label{alg:heuristic_search}
\begin{algorithmic}[1]

\State \textbf{Input:} Candidate blocks $\mathcal{B}$, timesteps $\mathcal{T}$,
evaluation functions $\texttt{EvalImg}$, $\texttt{EvalTxt}$, $\texttt{EvalFus}$

\State \textbf{Output:} Optimal fusion block-timestep pair
$((b_{\mathrm{img}}^\ast, t_{\mathrm{img}}^\ast),
(b_{\mathrm{txt}}^\ast, t_{\mathrm{txt}}^\ast))$

\Statex
\State \textbf{Step 1: Identify unimodal optima}
\State $(b_{\mathrm{img}}', t_{\mathrm{img}}')
\gets \arg\max\limits_{b_{\mathrm{img}} \in \mathcal{B},\, t_{\mathrm{img}} \in \mathcal{T}}
\texttt{EvalImg}(b_{\mathrm{img}}, t_{\mathrm{img}})$
\State $(b_{\mathrm{txt}}', t_{\mathrm{txt}}')
\gets \arg\max\limits_{b_{\mathrm{txt}} \in \mathcal{B},\, t_{\mathrm{txt}} \in \mathcal{T}}
\texttt{EvalTxt}(b_{\mathrm{txt}}, t_{\mathrm{txt}})$

\Statex
\State \textbf{Step 2: Construct local neighborhood}
\State $\mathcal{C} \gets \text{Neighbors}\bigl(
(b_{\mathrm{img}}', t_{\mathrm{img}}'),
(b_{\mathrm{txt}}', t_{\mathrm{txt}}')
\bigr)$ 

\Statex
\State \textbf{Step 3: Local fusion search}
\State $((b_{\mathrm{img}}^\ast, t_{\mathrm{img}}^\ast),
(b_{\mathrm{txt}}^\ast, t_{\mathrm{txt}}^\ast))
\gets
\arg\max\limits_{((b_{\mathrm{img}}, t_{\mathrm{img}}),
(b_{\mathrm{txt}}, t_{\mathrm{txt}})) \in \mathcal{C}}
\texttt{EvalFus}(b_{\mathrm{img}}, t_{\mathrm{img}},
                b_{\mathrm{txt}}, t_{\mathrm{txt}})
$

\State \Return $((b_{\mathrm{img}}^\ast, t_{\mathrm{img}}^\ast),
(b_{\mathrm{txt}}^\ast, t_{\mathrm{txt}}^\ast))$

\end{algorithmic}
\end{algorithm}

Let \(\mathbf{h}_{\mathrm{img}} \in \mathbb{R}^{d_{\mathrm{img}} \times 1}\) and \(\mathbf{h}_{\mathrm{txt}} \in \mathbb{R}^{d_{\mathrm{txt}} \times 1}\) denote the diffusion representations from image and language continuous diffusion pre-trained models, respectively, where \(d_{\mathrm{img}}\) and \(d_{\mathrm{txt}}\) represent the dimensionalities of image and text features, which need not be equal. We explore several strategies to combine them before feeding them into the multi-label classifier: (1) directly concatenating them, i.e., \(\text{Concat}(\mathbf{h}_{\mathrm{img}}, \mathbf{h}_{\mathrm{txt}})\); (2) firstly performing a linear projection to \(\mathbf{h}_{\mathrm{img}}\) and \(\mathbf{h}_{\mathrm{txt}}\), then concatenating them, i.e., \(\text{Concat}(\mathbf{W}_{\mathrm{img}}\mathbf{h}_{\mathrm{img}}, \mathbf{W}_{\mathrm{txt}}\mathbf{h}_{\mathrm{txt}})\), where \(\mathbf{W}_{\mathrm{img}} \in \mathbb{R}^{d_{\mathrm{alg}} \times d_{\mathrm{img}}}\), \(\mathbf{W}_{\mathrm{txt}} \in \mathbb{R}^{d_{\mathrm{alg}} \times d_{\mathrm{txt}}}\), and \(d_{\mathrm{alg}}\) denotes the dimensionality of the shared alignment space for image and text representations; (3)firstly performing a linear projection to \(\mathbf{h}_{\mathrm{img}}\) and \(\mathbf{h}_{\mathrm{txt}}\), then adding them, i.e., \(\mathbf{W}_{\mathrm{img}}\mathbf{h}_{\mathrm{img}} +\mathbf{W}_{\mathrm{txt}}\mathbf{h}_{\mathrm{txt}}\); (4) Cross attention: the image representations \(\mathbf{h}_{\mathrm{img}}\) are used as queries, while the text features \(\mathbf{h}_{\mathrm{txt}}\) serve as both keys and values, i.e., \(\text{CrossAttention}(\mathbf{h}_{\mathrm{img}}, \mathbf{h}_{\mathrm{txt}})=\operatorname{softmax}\!\Bigl(
      \frac{\mathbf{W}_Q\mathbf{h}_{\mathrm{img}}  (\mathbf{W}_K\mathbf{h}_{\mathrm{txt}} )^{\!\top}}{\sqrt{d_k}}
    \Bigr)
   \mathbf{W}_V \mathbf{h}_{\mathrm{txt}} \), where \(\mathbf{W}_Q \in \mathbb{R}^{d_k \times d_{\mathrm{img}}}, \mathbf{W}_K, \mathbf{W}_V \in \mathbb{R}^{d_k \times d_{\mathrm{txt}}}\) are learned projection matrices, and \(d_k\) is the key dimensionality.

\section{Experiments}
\label{sec: Experiments}

\textbf{Datasets.} We consider two primary multi-modal, multi-label datasets: VG500~\citep{VG500}, a widely used subset of Visual Genome~\citep{VGdataset}, and Pix2Cap-COCO~\citep{you2025pix2capcocoadvancingvisualcomprehension}. We aggregate the region-level captions within each image into a single textual description, which is used as the corresponding image-level caption in our experiments. We further evaluate our method on additional benchmark datasets, including MS-COCO (2014)~\citep{lin2014microsoft}, CIFAR-100~\citep{cifar}, Tiny-ImageNet~\citep{tinyimagenet}, and PASCAL VOC 2007~\citep{voc12}, to assess its robustness across different dataset scales and label configurations.

\textbf{Evaluation metrics.} We evaluate multi-label classification performance using a comprehensive set of metrics, including mean average precision (mAP), per-class precision (CP), per-class Recall (CR), per-class F1 (CF1), overall precision (OP), overall recall (OR) and overall F1 (OF1).

\textbf{Linear classifier setting.} The extracted diffusion features are evaluated using a single-layer linear classifier trained with Binary Cross-Entropy (BCE) loss for multi-label tasks or Cross-Entropy loss for single-label tasks. For the linear probing setting, the classifier is trained using the Adam optimizer with an initial learning rate of \(1 \times 10^{-3}\), together with a cosine annealing learning rate schedule over 40 epochs. A batch size of \(128\) is used by default unless stated otherwise.

\textbf{Experiments settings.} All experiments are conducted using distributed training (DDP) across \(4 \times\) NVIDIA GeForce RTX \(4090\) GPUs, with automatic mixed precision (AMP) enabled to accelerate training.

\subsection{Image-only diffusion representation}
\label{sec: image representation}
\textbf{Model architecture.} For the image modality, we utilize the latent space DiT model~\citep{Peebles_2023_ICCV} as the pre-trained backbone to extract image diffusion representations. We retrieve the DiT-XL/2 checkpoint, pretrained on \(256^2\) ImageNet from its official codebase for class-conditional generation. We employ it in an unconditional manner by setting the label to null~\citep{ho2021classifierfree}. DiT-XL/2 has \(28\) Transformer layers, a hidden size of \(1152\), and \(16\) attention heads, following the largest configuration of the DiT model family. The off-the-shelf VAE~\citep{kingma2014autoencoding} model for latent compression has a down-sample factor of \(8\), retrieved from Stable Diffusion~\citep{stablediffusion}. 

To systematically analyze the behavior of intermediate activations in image diffusion models, we perform ablation studies by extracting features from different diffusion timesteps and Transformer blocks, and evaluating their performance on multi-label classification under the linear probing setting.

\begin{figure}[htbp]
  \centering
  \includegraphics[width=0.5\textwidth]{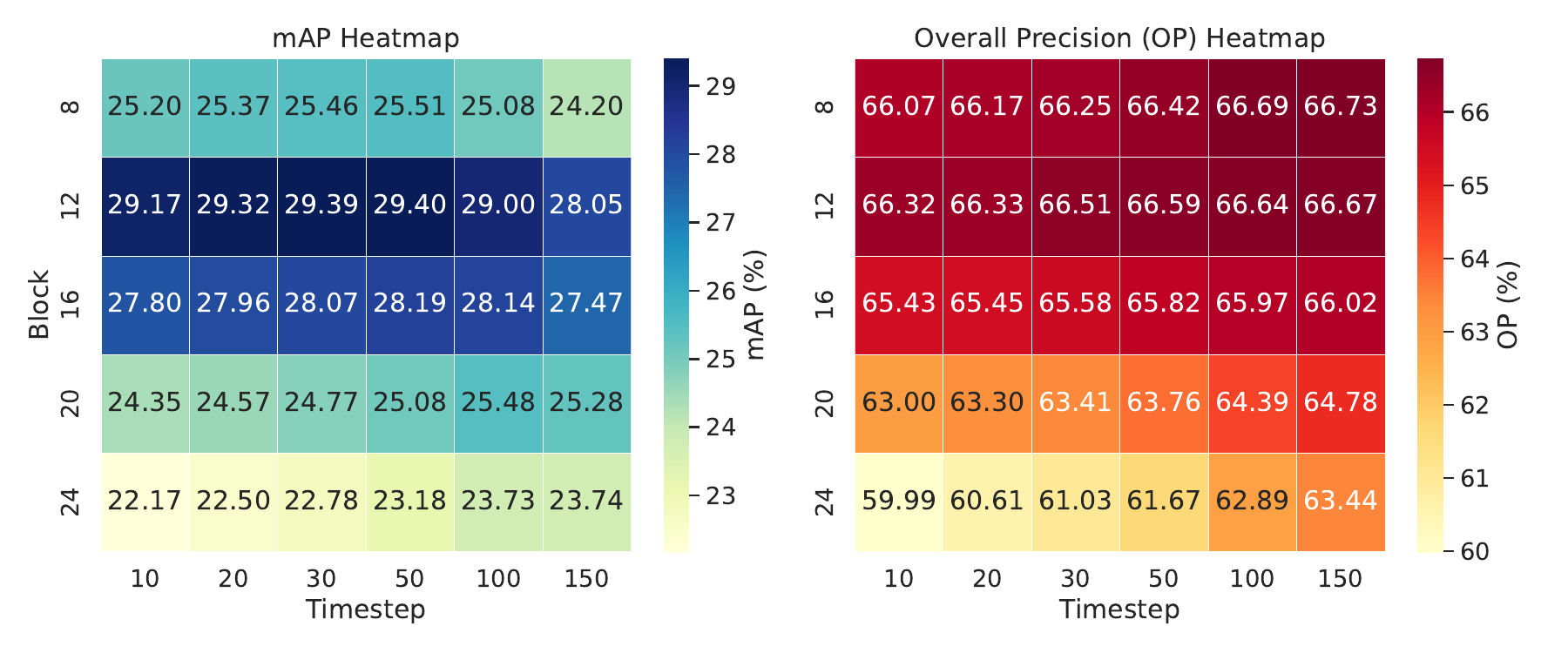}
  \caption{Multi-label classification performance of image diffusion representations across different Transformer blocks and diffusion timesteps on the VG500. Brighter regions indicate higher mAP (left) or OP (right) values, highlighting the optimal selection of features.}
  \label{fig:image_map_heatmap}
\end{figure}

As shown in Fig.~\ref{fig:image_map_heatmap}, classification performance measured by mAP and OP exhibits a consistent unimodal trend along the diffusion timestep dimension, where intermediate timesteps yield superior results compared to early or heavily corrupted states.

Along the depth dimension, we observe a highly consistent pattern across timesteps: features extracted from the 12th Transformer block consistently achieve the best performance compared to other layers. This empirical result suggests that intermediate layers of the diffusion-Transformer strike a favorable balance between semantic abstraction and spatial specificity, making them particularly well suited for multi-label classification.

\subsection{Text-only diffusion representation}
\label{sec: text representation}

\textbf{Model architecture.} 
For the text modality, we employ the Plaid \(1\mathrm{B}\) model~\citep{plaid} as a pre-trained language diffusion backbone to extract diffusion-based text representations. Plaid \(1\mathrm{B}\) is a Transformer-based model with approximately \(1.3\) billion parameters, whose denoising network consists of \(24\) Transformer blocks with a hidden dimension of \(2048\).

For consistency across experiments, the input sequence length is fixed to \(600\) tokens on VG500 and \(150\) tokens on Pix2Cap-COCO.

\begin{figure}[htbp]
  \centering
  \includegraphics[width=\linewidth]{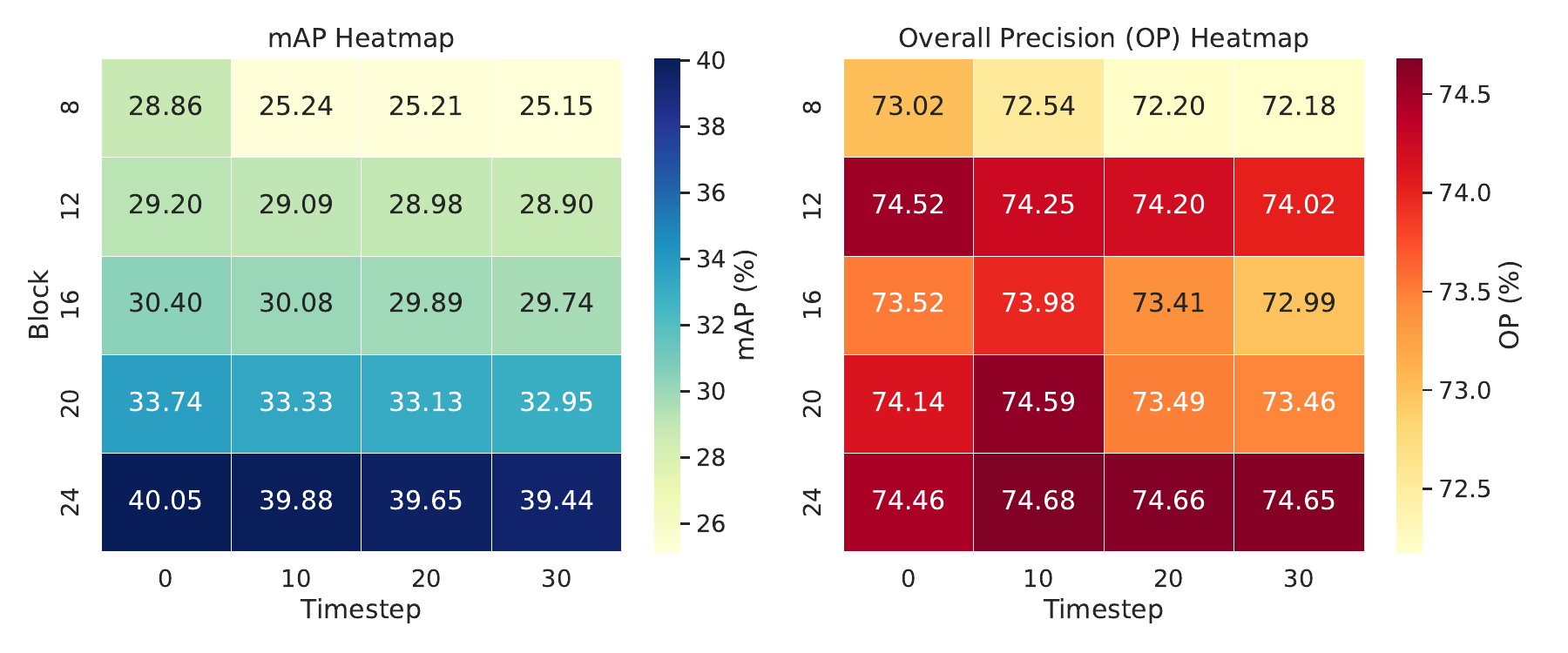}
  \caption{
  Multi-label classification performance of text diffusion representations across different Transformer blocks and diffusion timesteps on the VG500. Brighter regions indicate higher mAP (left) or OP (right) values.}
  \label{fig:languag_block_timestep_heatmap}
\end{figure}

Following the same evaluation protocol as in the image branch, we perform multi-label classification using the extracted text diffusion representations. As illustrated in the fig.~\ref{fig:languag_block_timestep_heatmap}, representations extracted from deeper Transformer blocks (e.g., block \(24\)) consistently achieve better performance across all diffusion timesteps. Moreover, the best performance is observed at the noise-free timestep (\(t=0\)), indicating that early diffusion steps preserve richer semantic information for text-based multi-label classification.

\subsection{Cross-modal fusion representation}

As discussed in Section~\ref{sec: representation fusion}, we investigate multiple fusion strategies to combine image and text diffusion representations for multi-label classification.

Specifically, we evaluate four feature fusion methods: \textit{Simple Concat}, \textit{Linear Concat}, \textit{Linear Addition}, and \textit{Cross Attention}. In all methods, image and text features are first \(\ell_2\)-normalized to ensure comparable feature scales across modalities. For \textit{Simple Concat}, the normalized features are directly concatenated. For \textit{Linear Concat}, \textit{Linear Addition}, and \textit{Cross Attention}, the normalized features are further projected into a shared embedding space via modality-specific linear layers before fusion. Unless otherwise specified, the dimensionality of the shared embedding space is set to \(d_{\mathrm{alg}} = d_k = 2048\).

\begin{figure}[htbp]
    \centering
    \includegraphics[width=\linewidth]{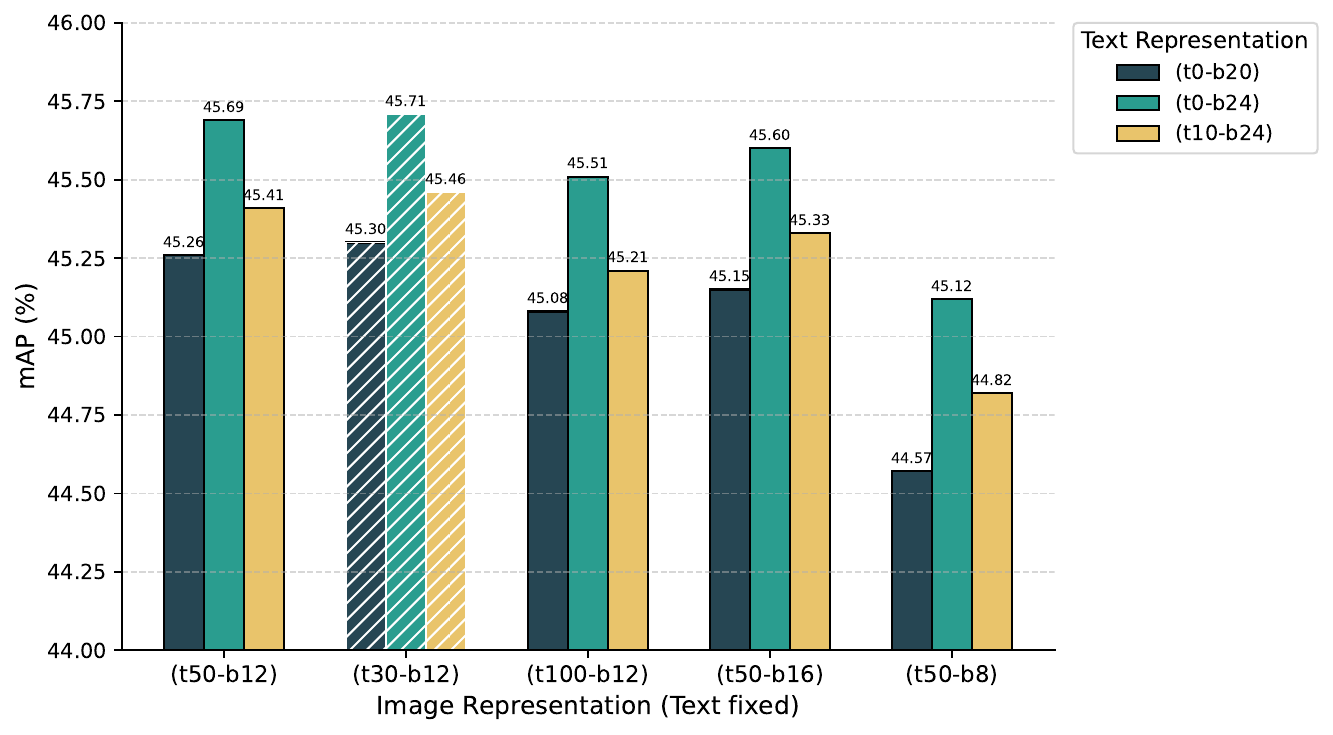}
    \caption{
        Multi-label classification performance (mAP) on VG500 using \textit{Linear Addition} fusion across different image–text representation pairs. Each group denotes an image representation (e.g., \texttt{(t50-b12)}), and each bar corresponds to a text representation. The best combination in each group is highlighted with white diagonal stripes.
    }
    \label{fig:image_text_fusion}
\end{figure}

Following the ablation study on fusion methods (Table~\ref{tab:fusion}), we evaluate fusion performance using the \textit{Linear Addition} method with the heuristic block-timestep selection strategy introduced in Section~\ref{sec: representation fusion}. As shown in Fig.~\ref{fig:image_text_fusion}, the best performance on VG500 is obtained by fusing image representations extracted at \(t_{\mathrm{img}} = 30, b_{\mathrm{img}} = 12\) with text representations extracted at \(t_{\mathrm{txt}} = 0, b_{\mathrm{txt}} = 24\), achieving a mAP of \(45.71\%\).

\begin{table}[htbp]
\centering
\caption{Ablation study of fusion methods on Pix2Cap-COCO (\(t_{\mathrm{img}}=50, b_{\mathrm{img}}=12; t_{\mathrm{txt}} = 0, b_{\mathrm{txt}} = 24\)).}
\label{tab:fusion}
\setlength{\tabcolsep}{2pt}
\begin{tabular}{lccccccc}
\toprule
\textbf{Fusion Method} & \textbf{mAP} & \textbf{CP} & \textbf{CR} & \textbf{CF1} & \textbf{OP} & \textbf{OR} & \textbf{OF1} \\
\midrule
Simple Concat & 78.0 & \textbf{85.5 }& 59.9 & 69.5 & \textbf{91.6 }& 67.8 & 77.91 \\
Linear Concat & 79.0 & 82.8 & 68.0 & 74.2 & 89.0 & 74.9 & 81.3 \\

Cross Attention & 76.8 & 83.6 & 64.9 & 72.5 & 89.9 & 72.1 & 80.0 \\
\textbf{Linear Add} & \textbf{79.6} & 82.9 & \textbf{68.2} & \textbf{74.4} & 89.2 & \textbf{75.1} & \textbf{81.6} \\
\bottomrule
\end{tabular}
\end{table}

We report the multi-label classification metrics for the optimal block–timestep combinations using Linear Addition fusion strategy, benchmarked against strong baselines on the VG500 (Table~\ref{tab:vg500_comparison}) and Pix2Cap-COCO (Table~\ref{tab:Pix2Cap-COCO_comparison}), where \textbf{"LP"} denotes linear probing, i.e., training a linear classifier on frozen features, and \textbf{"FT"} denotes fine-tuning the image diffusion model for feature extraction, after which a classifier is trained on the resulting features.

For VG500, the classifier is first trained for 20 epochs with a learning rate of $1\times10^{-3}$, followed by joint fine-tuning of the classifier and diffusion model for 2 epochs with a learning rate of $1\times10^{-4}$. For Pix2Cap-COCO, the initial classifier training also uses 20 epochs with a learning rate of $1\times10^{-3}$, while joint fine-tuning is conducted for 5 epochs with a learning rate of $5\times10^{-5}$. In all fine-tuning experiments, the batch size is set to 16.

The optimal LP configuration on VG500 is $(t_{\text{img}}=30, b_{\text{img}}=12, t_{\text{txt}}=0, b_{\text{txt}}=24)$, while FT uses $(t_{\text{img}}=50, b_{\text{img}}=16, t_{\text{txt}}=0, b_{\text{txt}}=24)$. On Pix2Cap-COCO, the best LP setting is $(t_{\text{img}}=50, b_{\text{img}}=12, t_{\text{txt}}=0, b_{\text{txt}}=24)$, while FT uses $(t_{\text{img}}=50, b_{\text{img}}=16, t_{\text{txt}}=0, b_{\text{txt}}=24)$.

\begin{table}[htbp]
\centering
\caption{Comparison with strong baseline models on VG500.}
\label{tab:vg500_comparison}
\begin{tabular}{lc}
\toprule
\textbf{Method} & \textbf{mAP(\%)} \\
\midrule
ResNet-101~\citep{kaiming2016residual} & 30.9 \\
ResNet-SRN~\citep{zhu2017spatial} & 33.5 \\
SS-GRL~\citep{chen2019semantic} & 36.6 \\
C-Tran~\citep{C-Tran}  & 38.4 \\
DRGN~\citep{zhoumining2024} & 39.8\\
DATran~\citep{DATran} & 40.1\\
SADCL~\citep{Ma2023SemanticAwareDC} & 40.5\\
Q2L-TResL-22k~\citep{Query2Label} & 42.5 \\
\midrule
ResNet-50~\citep{kaiming2016residual} + LLaMA 3.2 1B~\citep{grattafiori2024llama3herdmodels} (LP) & 40.5 \\
CLIP (LP)~\citep{radford2021learningtransferablevisualmodels} & 42.9 \\
ViLT (LP)~\citep{kim2021viltvisionandlanguagetransformerconvolution} & 37.3 \\
\textbf{Diff-Feat (Ours, LP)} & 45.7\\
\textbf{Diff-Feat (Ours, FT)} & \textbf{46.1}\\
\bottomrule
\end{tabular}
\end{table}

\begin{table}[htbp]
\centering
\caption{Comparison with strong baseline models on Pix2Cap-COCO.}
\label{tab:Pix2Cap-COCO_comparison}
\setlength{\tabcolsep}{2pt}
\begin{tabular}{lccccccc}
\toprule
\textbf{Method} & \textbf{mAP} & \textbf{CP} & \textbf{CR} & \textbf{CF1} & \textbf{OP} & \textbf{OR} & \textbf{OF1} \\
\midrule
ResNet-50 + LLaMA3.2 1B (LP) & 70.4 & 73.9 & 58.1 & 64.6 & 82.3 & 65.9 & 73.2 \\
CLIP (LP) & 76.3 & \textbf{84.4} & 57.8 & 67.1 & \textbf{91.8} & 65.9 & 76.7 \\
ViLT (LP) & 73.0 & 84.0 & 56.9 & 66.6 & 90.7 & 64.0 & 75.1 \\
\midrule
\textbf{Diff-Feat (Ours, LP)} & 79.6 & 82.9 & \textbf{68.2} & 74.4 & 89.2 & \textbf{75.1} & \textbf{81.6} \\
\textbf{Diff-Feat (Ours, FT)} & \textbf{80.2} & 83.7 & 68.0 & \textbf{74.7} & 89.5 & 75.0 & \textbf{81.6} \\
\bottomrule
\end{tabular}
\end{table}

\subsection{Empirical observation: modality-specific trends in diffusion representations}
\label{sec:empirical_observation}

\textbf{Notation.} Let \(\mathbf{x}\) denote an input instance(either an image or text embedding). We apply a pre-trained diffusion model to \(\mathbf{x}\) via the forward process, yielding latent states \(\mathbf{z}_t\) at timestep \(t\) (through Eq.~\ref{eq:diffusion_forward}). Let \(\mathbf{h}_{t,b}\) denote the hidden representation extracted from the \(b\)-th Transformer block when \(\mathbf{z}_t\) is input. Define \(\mathcal{A}(t, b)\) as downstream tasks performance (e.g., mAP) using \(\mathbf{h}_{t,b}\) as features for linear probing.
Empirical findings reveal modality-specific trends in \(\mathcal{A}(t, b)\) (Fig.~\ref{fig:image_map_heatmap} and~\ref{fig:languag_block_timestep_heatmap}):

\begin{itemize}
    \item[\textbf{(1)}] \textbf{Image modality, fixed \(b\):} \(\mathcal{A}(t,b)\) is unimodal in \(t\); there exists \(t^*(b)\) such that
    \[
    \mathcal{A}(t, b)\text{ increases for } t < t^*(b), \quad \text{decreases for } t > t^*(b).
    \]
    
    \item[\textbf{(2)}] \textbf{Image modality, fixed \(t\):} \(\mathcal{A}(t,b)\) is unimodal in \(b\); there exists \(b^*(t)\) such that
    \[
    \mathcal{A}(t, b)\text{ increases for } b < b^*(t), \quad \text{decreases for } b > b^*(t).
    \]
    
    \item[\textbf{(3)}] \textbf{Text modality, fixed \(b\):} \(\mathcal{A}(t,b)\) decreases monotonically with \(t \in \mathcal{T}\).
    
    \item[\textbf{(4)}] \textbf{Text modality, fixed \(t\):} \(\mathcal{A}(t,b)\) increases monotonically with \(b \in \mathcal{B}\).
\end{itemize}

\begin{figure*}[t]
    \centering

    \begin{subfigure}[t]{0.4\textwidth}
        \centering
        \includegraphics[width=\linewidth]{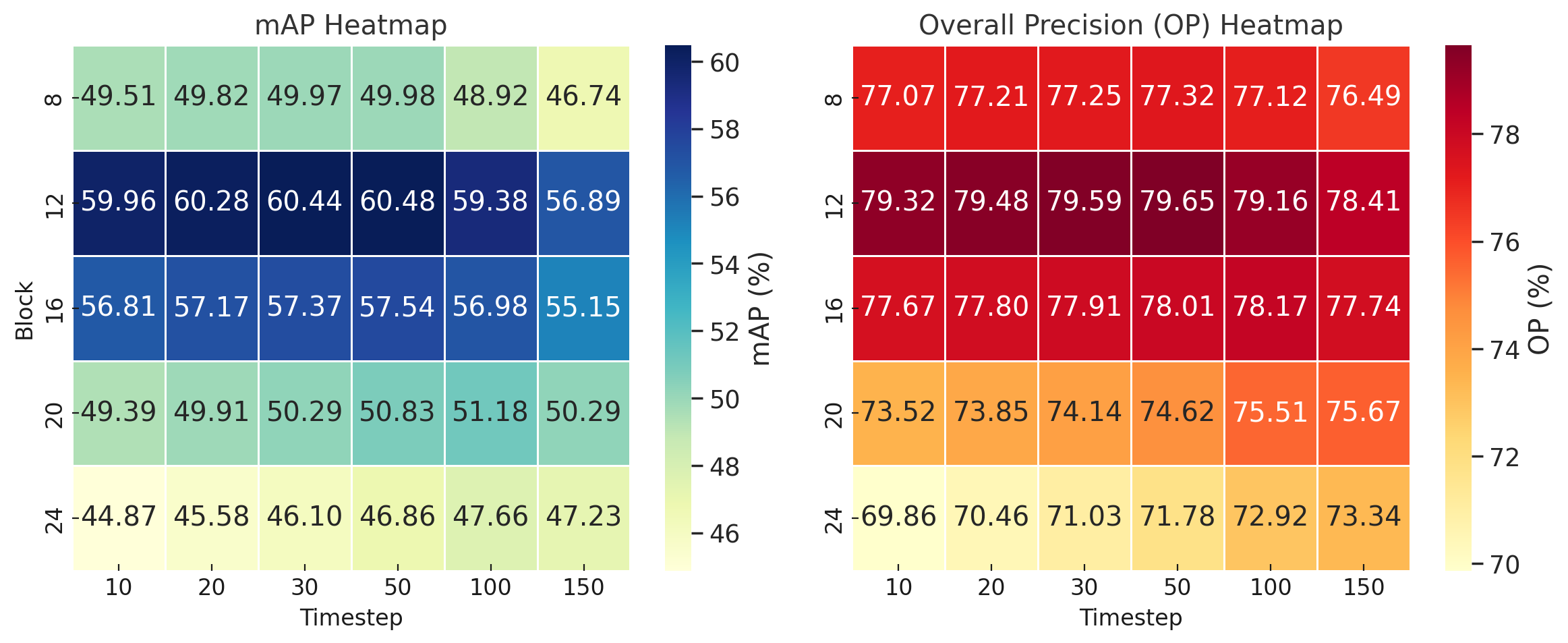}
        \caption{MS-COCO}
    \end{subfigure}
    \hspace{0.01\textwidth}
    \begin{subfigure}[t]{0.4\textwidth}
        \centering
        \includegraphics[width=\linewidth]{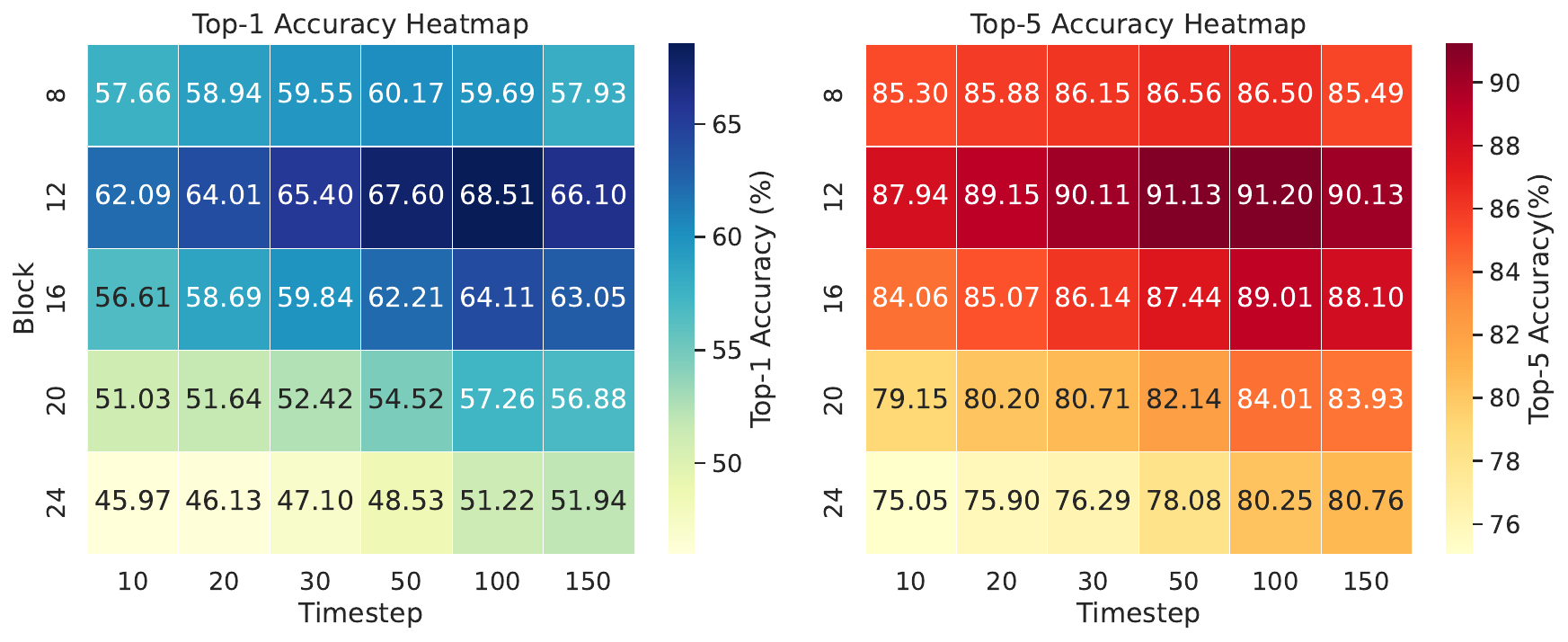}
        \caption{CIFAR-100}
    \end{subfigure}

    \vspace{0.4em}

    \begin{subfigure}[t]{0.4\textwidth}
        \centering
        \includegraphics[width=\linewidth]{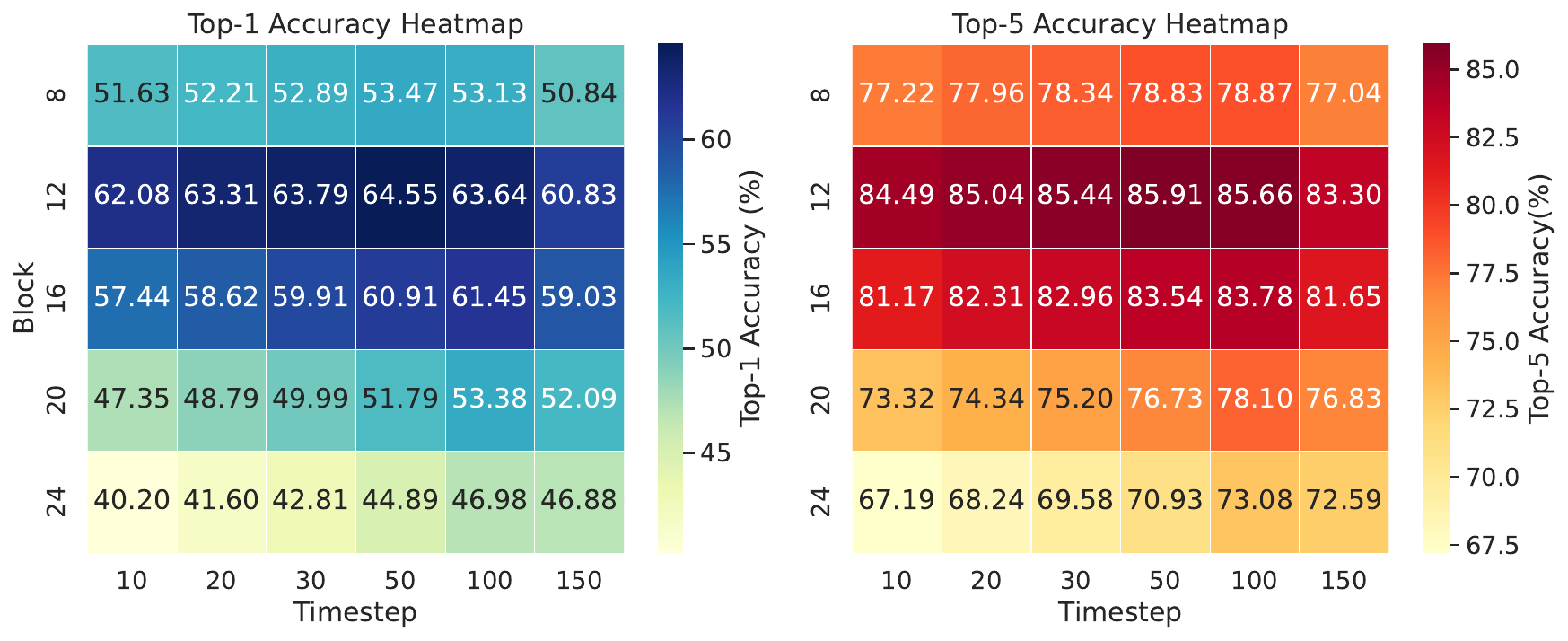}
        \caption{Tiny-ImageNet}
    \end{subfigure}
    \hspace{0.01\textwidth}
    \begin{subfigure}[t]{0.4\textwidth}
        \centering
        \includegraphics[width=\linewidth]{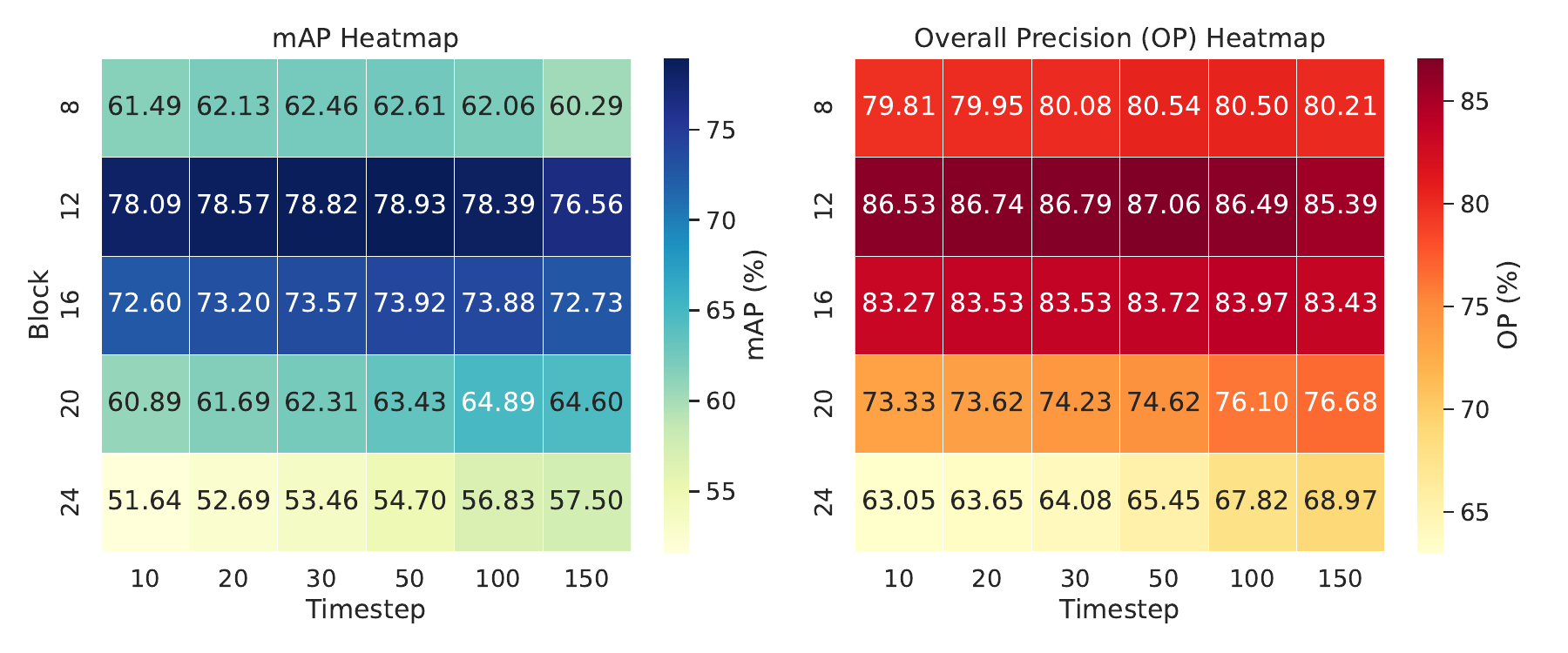}
        \caption{PASCAL VOC 2007}
    \end{subfigure}

    \caption{Heatmap visualizations of classification performance across diffusion timesteps and Transformer blocks on multiple image benchmarks.}
    \label{fig:heatmap_all_datasets}
\end{figure*}

As shown in Fig.~\ref{fig:heatmap_all_datasets}, for image data, the optimal diffusion timestep tends to vary across a broad range, with local fluctuations. In contrast, the optimal Transformer block is consistently fixed at \textbf{layer 12}. Although this may partially relate to the DiT model architecture (which contains \(28\) layers), it appears remarkably invariant across downstream tasks, dataset distributions, and evaluation metrics. This consistent pattern may offer insights into the internal workings of black-box diffusion models, and motivates the subsequent experimental and mechanism-level analyses presented in Sections~\ref{sec: scaling} and~\ref{sec: understanding}.

\subsection{Scaling Analysis of Optimal Layer Position in Vision Diffusion Models}
\label{sec: scaling}

We further evaluate diffusion-Transformer models with different architectural configurations (U-ViT~\citep{bao2022all} variants) to assess the robustness of the "\textbf{Middle Layer}" phenomenon. As shown in Table~\ref{tab:optimal_layer}, similar trends persist across model depths, with optimal performance consistently occurring in intermediate layers, demonstrating that this behavior generalizes across architectures. Moreover, we observe a \textit{strong positive correlation} between the optimal layer index and the total number of layers (Spearman \(\rho=0.97\), \(p<0.0015\)), indicating that deeper models tend to shift the optimal layer proportionally deeper.

In the U-ViT family, “S”, “L”, and “H” denote small, large, and huge variants, respectively, while “/2” and “/4” indicate different input patch size. The “Deep” variant refers to models with increased Transformer depth.
\begin{table}[htbp]
\centering
\setlength{\tabcolsep}{2pt}
\caption{Optimal Layer Selection Across Different Architectures at Diffusion Timestep $t=50$ on the MS-COCO dataset.}
\label{tab:optimal_layer}
\begin{tabular}{lccc}
\toprule
\textbf{Model Architecture} & \textbf{Optimal Layer} & \textbf{Total Layers} & \textbf{mAP}\\
\midrule
CIFAR10 (U-ViT-S/2) & 10 & 13 & 28.24\\
CelebA 64x64 (U-ViT-S/4) & 9 & 13 & 27.64\\
ImageNet 256x256 (U-ViT-L/2)   & 12   &  21 & 59.15 \\
ImageNet 512x512 (U-ViT-L/4)   & 11   &  21 & 54.88\\
ImageNet 256x256 (U-ViT-H/2)   & 16   &  29 & 64.32\\
ImageNet 512x512 (U-ViT-H/4)   & 16   &  29 & 61.16 \\

\bottomrule
\end{tabular}
\end{table}

\subsection{Understanding Modality-Dependent Optimality in Diffusion Models}
\label{sec: understanding}

\textbf{Vision: semantic granularity versus noise robustance.}
To investigate how semantic granularity manifests in the diffusion representation, we analyze robustness along two orthogonal axes: the diffusion timestep $t$ and the Transformer block depth~$b$. For the timestep axis, we normalize each task’s performance by its peak value \(r_{\text{task}}(t)=
\frac{\mathrm{mAP}_{\text{task}}(t)}{\max_t \mathrm{mAP}_{\text{task}}(t)}
\), and define the \textit{$\alpha$-retention interval} as \(
\mathcal{T}_{\alpha}(\text{task})
=
\{t \mid r_{\text{task}}(t)\ge \alpha \}
\), where $\alpha=0.95$ denotes the range of diffusion timesteps that preserve at least 95\% of the task’s peak mAP.  
Along the depth axis, an analogous normalization is applied over Transformer blocks,
$r_{\text{task}}(b)=\mathrm{mAP}_{\text{task}}(b)/\max_b \mathrm{mAP}_{\text{task}}(b)$,
allowing us to compare how coarse- and fine-grained semantics emerge across layers.

Experiments are conducted on MS-COCO using object labels (fine-grained) and super-category labels (coarse-grained), with linear probes trained on representations extracted from block $b=12$ of DiT-XL/2-256$\times$256.

\begin{figure}[htbp]
  \centering
  \begin{minipage}[t]{0.22\textwidth}
    \centering
    \includegraphics[width=\textwidth]{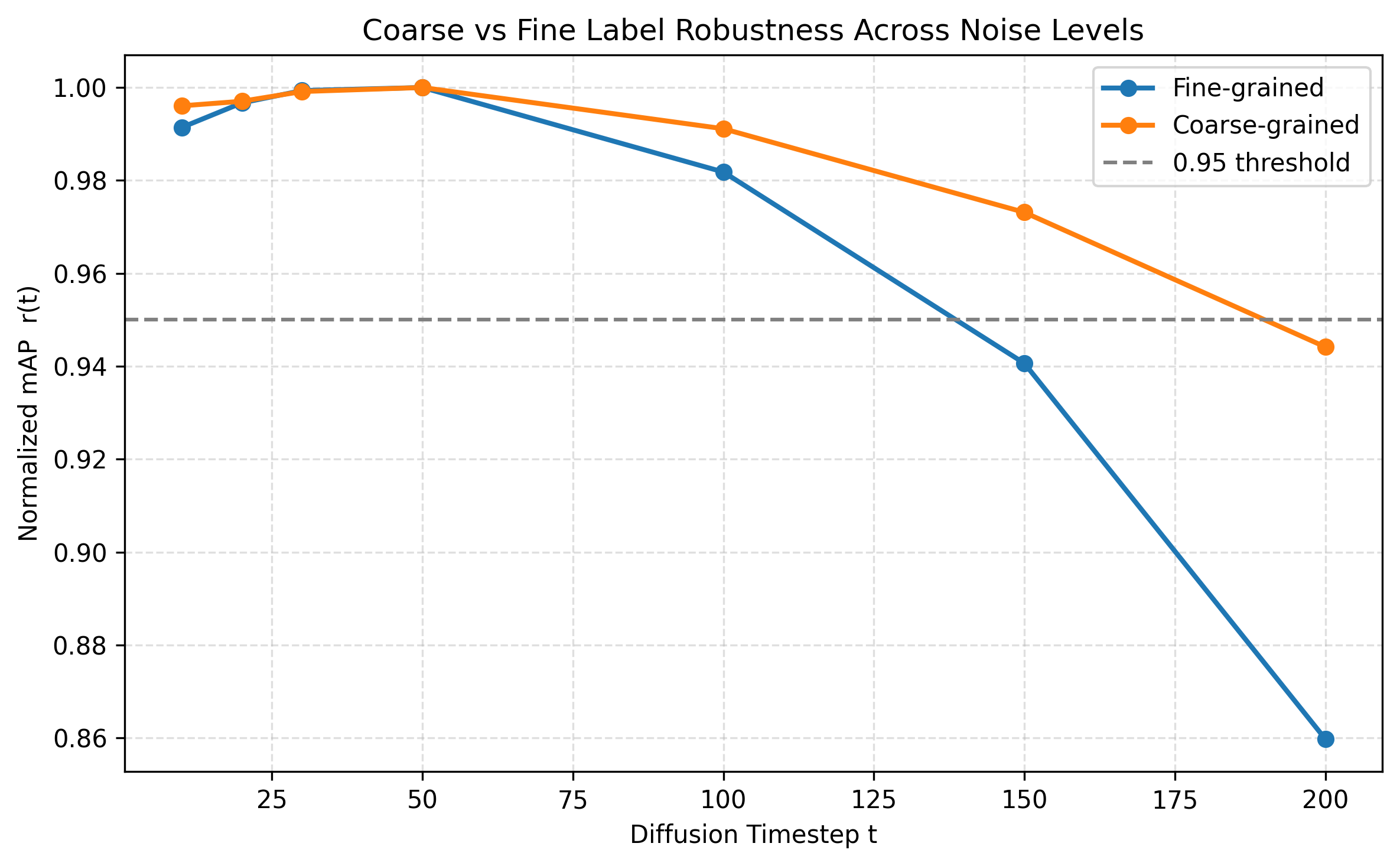}
    \caption*{(a) Fixed block $b=12$.}
  \end{minipage}
  \hfill
  \begin{minipage}[t]{0.22\textwidth}
    \centering
    \includegraphics[width=\textwidth]{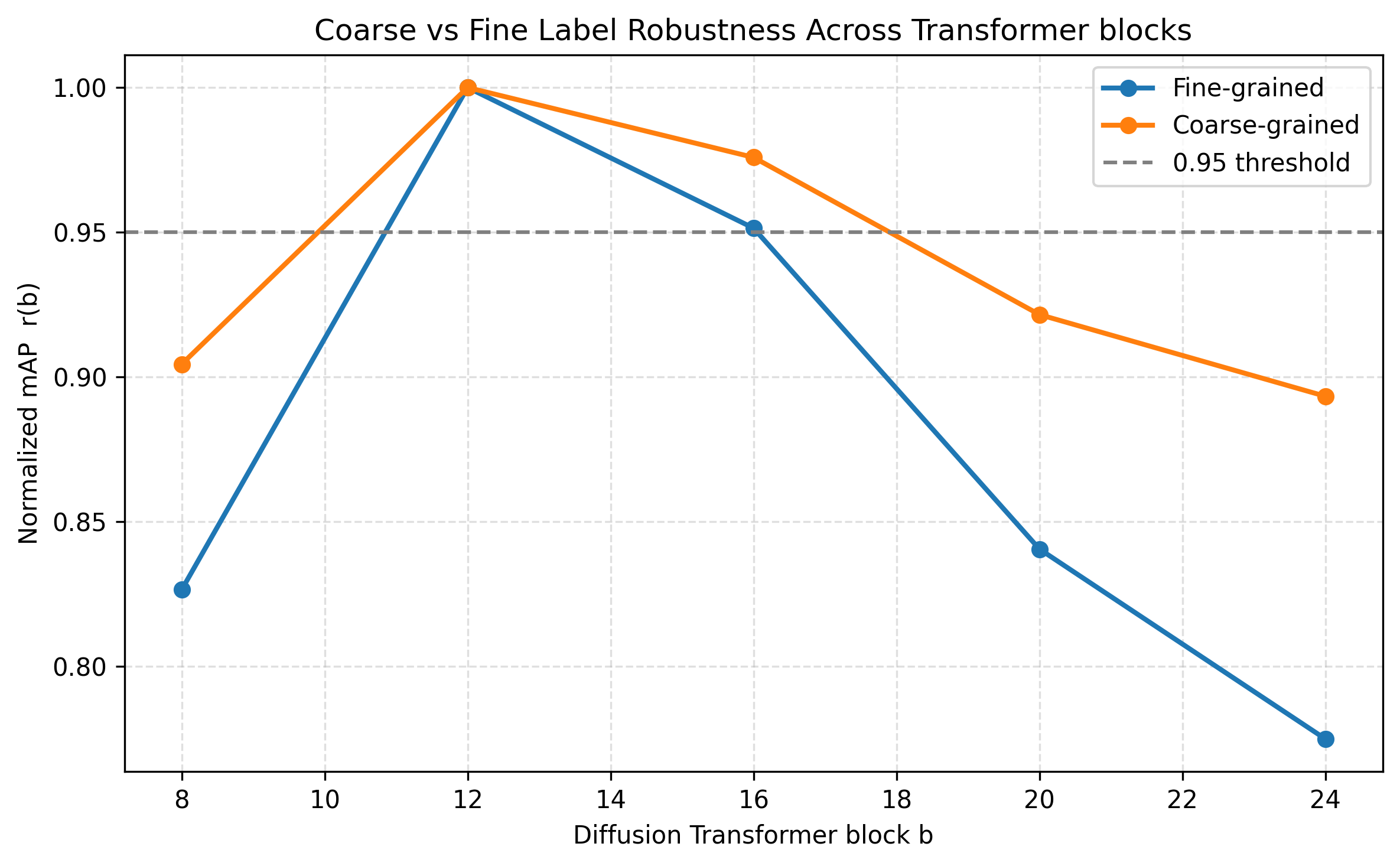}
    \caption*{(b) Fixed diffusion timestep $t=50$.}
  \end{minipage}

  \caption{
    Robustness comparison between coarse- and fine-grained labels along two orthogonal axes of DiT on the MS-COCO dataset.
  }
  \label{fig:coarse_fine_axes}
\end{figure}

As shown in Fig.~\ref{fig:coarse_fine_axes}, the fine-grained task drops below the 0.95 threshold at $t=150$ ($r_{\text{fine}}(150)=0.941$), yielding
$[10,150]\ge \mathcal{T}_{0.95}(\text{fine}) \ge [10,100]$.
In contrast, the coarse-grained task remains above the threshold even under strong noise corruption ($r_{\text{coarse}}(150)=0.973$), yielding
$[10,300]\ge \mathcal{T}_{0.95}(\text{coarse}) \ge [10,200]$.
Thus,
\(|\mathcal{T}_{0.95}(\text{coarse})|
\gg
|\mathcal{T}_{0.95}(\text{fine})|.
\)

A complementary pattern emerges along the \textit{depth} axis: when fixing $t$ and sweeping over blocks $b$, fine-grained labels exhibit a sharp performance peak in the mid-to-late layers (e.g., around block 12), whereas coarse-grained labels remain comparatively stable across a broader span of blocks. Together, the $t$-axis and $b$-axis analyses reveal a consistent semantic hierarchy: fine-grained labels rely on detail-sensitive, higher-frequency representations that emerge in later blocks and are easily disrupted by noise, while coarse-grained labels rely primarily on lower-frequency structural cues that persist under strong corruption and across many layers.

\textbf{Language: semantic neighborhood degradation under noise perturbation.}
To quantify how diffusion noise affects the local semantic structure of text embeddings, we analyze neighborhood relationships across diffusion timesteps at a fixed Transformer block depth.
Let $h(x_t^i)\in\mathbb{R}^{d_{\text{txt}}}$ denote the embedding of sample $x^i$ at diffusion timestep $t$, and let
$
N_t(i)=\mathrm{kNN}(h(x_t^i))
$
denote the set of $k$ nearest neighbors under cosine similarity, where $k=10$.

We consider two complementary neighborhood-based metrics to characterize geometric and semantic stability.
\emph{Neighborhood overlap} measures geometric stability and is defined as
$
\mathrm{Overlap}(t)=\sum_{i=1}^n\big[|N_0(i)\cap N_t(i)|/k\big].
$
\emph{Label consistency} measures semantic alignment and is defined as
$
\mathrm{Consistency}(t)=\sum_{i=1}^n\big[\frac{1}{k}\sum_{j\in N_t(i)}\mathbf{1}(y_i\cap y_j\neq\emptyset)\big],
$
where $y_i$ denotes the multi-label annotation associated with sample $i$.

\begin{figure}[htbp]
  \centering
  \begin{minipage}[t]{0.22\textwidth}
    \centering
    \includegraphics[width=\textwidth]{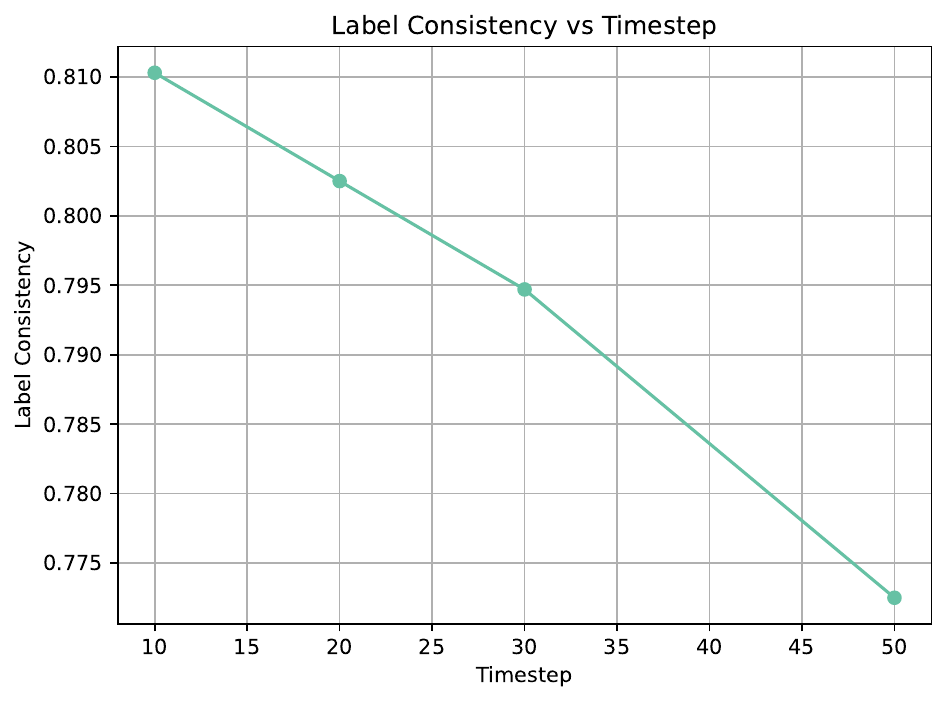}
    \caption*{(a) Label consistency.}
  \end{minipage}
  \hfill
  \begin{minipage}[t]{0.22\textwidth}
    \centering
    \includegraphics[width=\textwidth]{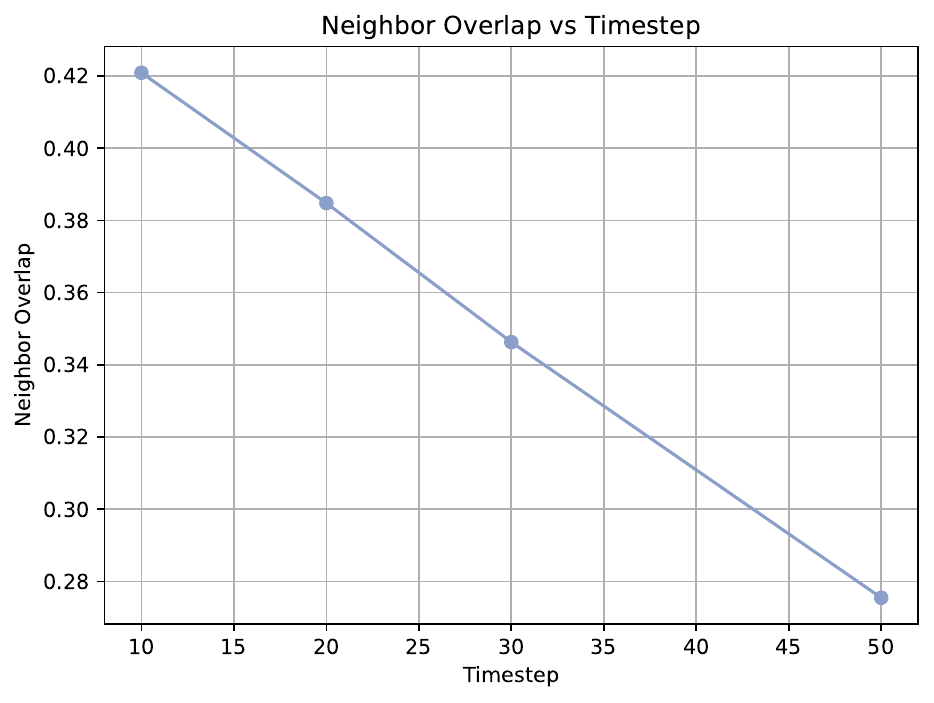}
    \caption*{(b) Neighborhood overlap.}
  \end{minipage}

  \caption{
    Semantic stability of text embeddings under different diffusion timestep at a fixed block depth $b=24$ on the VG500.
  }
  \label{fig:text_mechinsm}
\end{figure}

Fig.~\ref{fig:text_mechinsm} presents both metrics as functions of the diffusion timestep $t$ at block depth $b=24$ of Plaid-1B.
As $t$ increases, neighborhood overlap decreases monotonically, indicating a progressive disruption of the local embedding geometry.
Label consistency also declines smoothly, reflecting a gradual loss of semantic coherence among nearest neighbors.
Together, these results demonstrate that diffusion noise systematically degrades both geometric and semantic neighborhood structure of text embeddings, suggesting that, in the textual domain, noise acts as a destructive perturbation rather than a beneficial regularizer.

\section{Discussion}

For images, adaptive noise levels and Transformer blocks help remove redundant details while preserving task-relevant features, leading to a peak in discriminative quality at the intermediate noise level and block~\citep{kim2024textitreveliointerpretingleveragingsemantic}; In contrast, text representations are more sensitive to corruption: once corrupted by noise, semantic information becomes difficult to recover~\citep{plaid}, resulting in a monotonic degradation with increasing noise level. Meanwhile, deeper transformer blocks use self-attention to better understand text context~\citep{transformer}, explaining the upward trend in block depth.

\begin{figure}[htbp]
    \centering
    \begin{subfigure}[b]{0.45\linewidth}
        \centering
        \includegraphics[width=\linewidth]{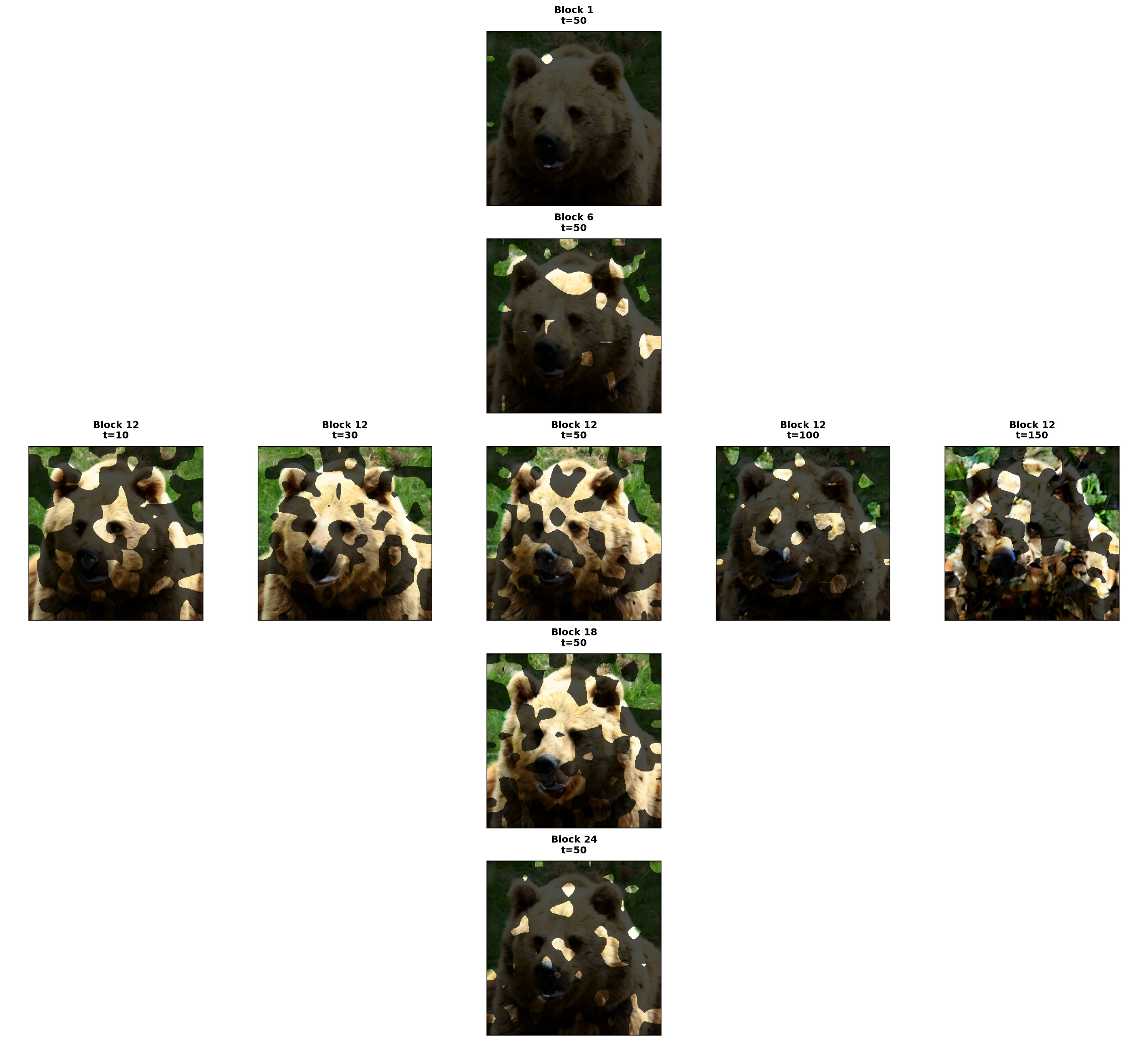}
        \label{fig:attn_map_bear}
    \end{subfigure}
    \hfill
    \begin{subfigure}[b]{0.45\linewidth}
        \centering
        \includegraphics[width=\linewidth]{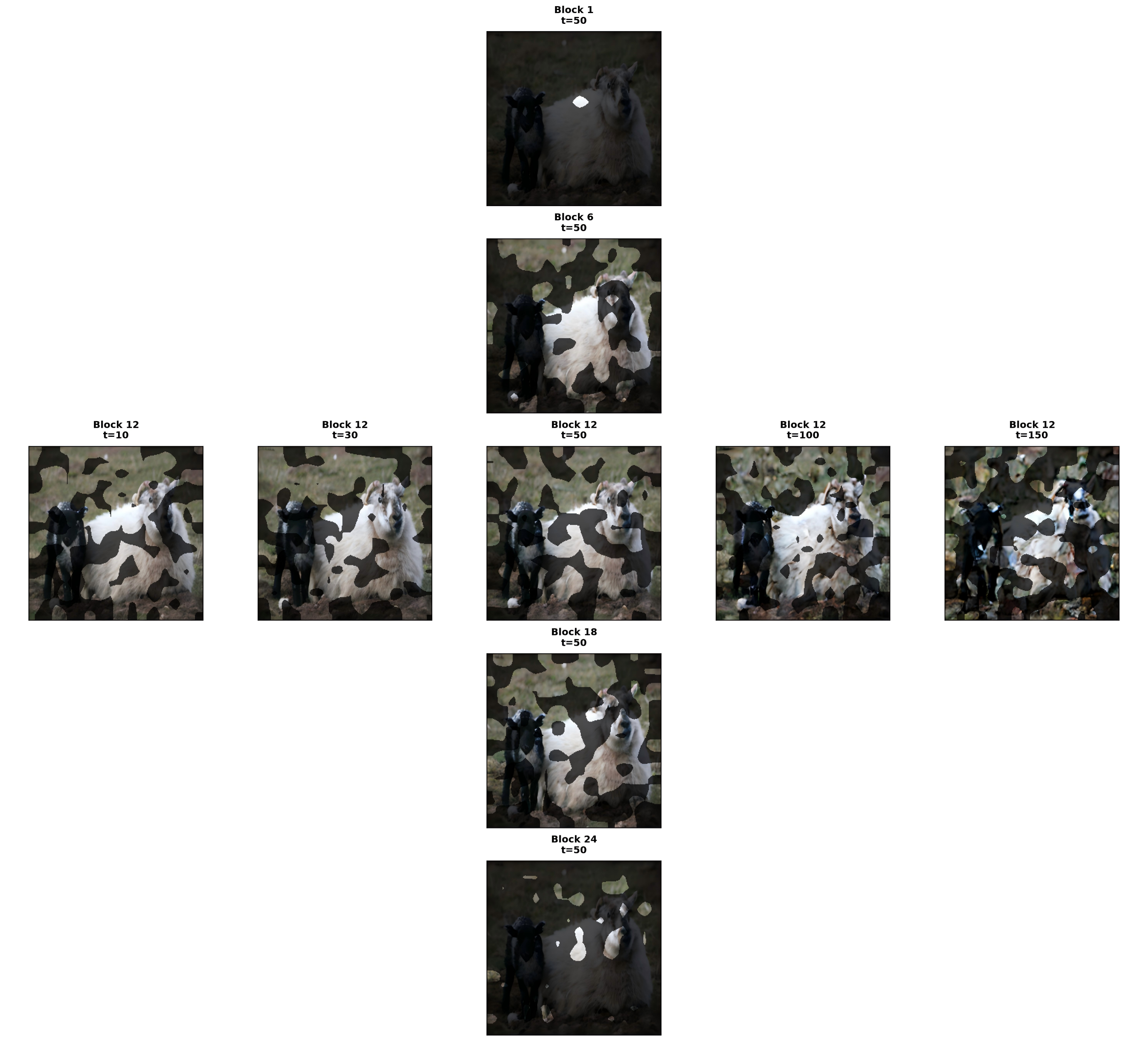}
        \label{fig:attn_map_sheep}
    \end{subfigure}
    \caption{
    Attention map visualization across diffusion timesteps and Transformer blocks for two representative samples in the MS-COCO.
    The maps demonstrate how the model gradually focuses on semantically meaningful regions.
    }
    
    \label{fig:attn_map}
\end{figure}

To further analyze the evolution of image representations across diffusion timesteps and Transformer layers, we visualize attention maps from the image diffusion model. Fig.~\ref{fig:attn_map} shows that intermediate timesteps and mid-level blocks yield more coherent and semantically focused attention, providing qualitative support for our quantitative findings.

Overall, our findings suggest that intermediate diffusion representations-properly selected in terms of block depth and diffusion timestep-can serve as strong and general-purpose features for downstream multi-label classification tasks.

We attribute the strong performance of the fused representation in highly imbalanced multi-label tasks to the powerful generative capacity of pre-trained diffusion models. In the meantime, the choice of optimal block-timestep pairs and effective fusion strategies plays a crucial role.

\section{Conclusions and future work}
\label{sec: conclusions}

In this paper, we propose \textit{Diff-Feat}, a simple yet effective framework for multi-label classification. By identifying optimal block-timestep combinations from image and text diffusion representations and leveraging a heuristic search strategy with a lightweight fusion mechanism, our method achieves strong performance, reaching \(46.1\%\) mAP on VG500 and \(80.2\%\) mAP on Pix2Cap-COCO, outperforming competitive baseline models. In addition, we provide new insights and mechanism-level analyses into the varying effectiveness of different block-timestep configurations across downstream tasks. Meanwhile, the scaling analysis further deepens our understanding of the representation properties of diffusion-Transformer models across different architectures, revealing the stability and consistency of optimal intermediate representations as model depth varies. We believe that \textit{Diff-Feat} offers a generalizable and adaptable solution for a wide range of multi-label classification scenarios, with potential applications as a strong self-supervised foundation in medical diagnosis and other specialized domains.

Despite its strong empirical performance, \textit{Diff-Feat} has several limitations. First, although the heuristic search strategy significantly reduces computational cost, it may overlook globally optimal fusion configurations, particularly in more complex settings. Second, as research on diffusion-based language models continues to advance, discrete diffusion language models have demonstrated strong generative performance~\citep{nie2025large}. However, the present work is limited to continuous diffusion language models, which constitutes an important direction for future research. Moreover, the proposed framework has the potential to be extended to real-world applications such as medical diagnosis, where it may contribute to positive societal impact and provide practical value in high-stakes decision-making scenarios.

\section*{Acknowledgments}
This work was supported by the MOE Project of Key Research Institute of Humanities and Social Sciences (22JJD110001).
\begingroup
\scriptsize
\setlength{\itemsep}{0pt}
\setlength{\parskip}{0pt}
\setlength{\parsep}{0pt}
\bibliographystyle{unsrt}
\bibliography{references}
\endgroup

\end{document}